\documentclass[final]{cvpr}

\usepackage{titling}
\usepackage{cvprtitle} 

\makeatletter
\@namedef{ver@everyshi.sty}{}
\makeatother

\usepackage{times}
\usepackage{epsfig}
\usepackage{graphicx}
\usepackage{glossaries} 
\usepackage{amsmath}
\usepackage{amssymb}
\usepackage{bm}
\usepackage{multibib}
\usepackage{caption}
\usepackage{subcaption}
\usepackage{tabularx}
\usepackage{color}
\usepackage{siunitx}

\usepackage{booktabs}

\definecolor{mygreen}{rgb}{0.2, 0.7, 0.1}

\usepackage[pagebackref=true,breaklinks=true,colorlinks,citecolor={mygreen},bookmarks=false]{hyperref}

\usepackage[capitalize]{cleveref}
\crefname{section}{Sec.}{Section}

\newacronym{iou}{IoU}{Intersection over Union}
\newacronym{cbn}{CBN}{Conditional Batch Normalization}
\newacronym{cnn}{CNN}{Convolutional Neural Network}
\newacronym{onet}{O-Net}{Occupancy Network}
\newacronym{mise}{MISE}{Multiresolution IsoSurface Extraction}
\newacronym{gpu}{GPU}{Graphics Processing Unit}
\newacronym{bb}{BB}{Bounding Box}
\newacronym{vram}{VRAM}{Video RAM}
\newacronym{lidar}{LiDAR}{Light Detection and Ranging}
\newacronym{fifo}{FIFO}{First In, First Out}

\DeclareSIUnit{\nothing}{\relax}

\newlength\figH
\newlength\figW
\newcommand{\Tensor}[1]{\boldsymbol{\mathrm{\MakeUppercase{#1}}}}

\newcommand{\Mean}[1]{\mathbb{E}\left(#1\right)}
\newcommand{\Var}[1]{\mathrm{Var}\left(#1\right)}

\usepackage[format=plain,labelformat=simple,labelsep=period,font=small,skip=4pt,compatibility=false]{caption}
\usepackage[font=footnotesize,skip=2pt,subrefformat=parens]{subcaption}

\usepackage{pifont} 
\newcommand{\cmark}{\textcolor{green}{\ding{51}}}%
\newcommand{\xmark}{\textcolor{red}{\ding{55}}}%

\usepackage{xspace}

\newcites{supp}{References}

\begin{document}

\title{Scalable 3D Semantic Segmentation for Gun Detection in CT Scans}

\author{Marius Memmel\thanks{equal contribution} \hspace{0.65cm} Christoph Reich\footnotemark[1] \hspace{0.65cm} Nicolas Wagner\footnotemark[1] \hspace{0.65cm} Faraz Saeedan
\\Department of Computer Science, TU Darmstadt\\{\tt\small \{marius.memmel, christoph.reich, nicolas.wagner\}@stud.tu-darmstadt.com}}

\maketitle

\begin{abstract}
With the increased availability of 3D data, the need for solutions processing those also increased rapidly. However, adding dimension to already reliably accurate 2D approaches leads to immense memory consumption and higher computational complexity. These issues cause current hardware to reach its limitations, with most methods forced to reduce the input resolution drastically. Our main contribution is a novel deep 3D semantic segmentation method for gun detection in baggage CT scans that enables fast training and low video memory consumption for high-resolution voxelized volumes. We introduce a moving pyramid approach that utilizes multiple forward passes at inference time for segmenting an instance.
\end{abstract}

\section{Introduction}
In the past decade, deep \glspl{cnn} have emerged as one of the most successful tools in computer vision~\cite{Yamashita2018,Khan2019}. For many 2D vision tasks, such as image classification~\cite{resnet, Huang2017, Dosovitskiy2021}, object detection~\cite{Bansal2020,Shen,Ren2015}, semantic segmentation~\cite{Chen2018, Ulku2019, MemmelGM2021, Prangemeier2021}, or optical flow estimation \cite{hur2020, Sun2018, Teed2020, Stone2021}, deep learning approaches yield reliable results. Transferring known 2D deep learning solutions to 3D volumes poses various difficulties. Most notably, the increased computational complexity to process 3D volumes, as well as their immense memory consumption, is not manageable by consumers or even state-of-the-art \glspl{gpu}~\cite{Shen}.\\
\indent Application areas reach from medical imaging over construction and planning to airport security. To ensure safety past the check-in, airport scanners are deployed to check passenger luggage before entering the restricted areas. A major goal of these is detecting prohibited items at the airport security screening. In our work, we will specifically focus on firearms. For this purpose, a private security company did create a dataset with 3D CT (computed tomography) scans of hand luggage containing the aforementioned items. Processing those scans on current \glspl{gpu} leads to excessive memory consumption culminating in small batch sizes and long loading times.\\
\indent In the following, we propose a novel 3D semantic segmentation method, the HiLo-Network, to overcome those challenges for this specific dataset. We assume that HiLo-Networks can be applied to a broader range of 3D datasets. The main goal is to reduce memory consumption while retaining a fast training process. We tradeoff multiple forward passes at inference time to obtain a scalable approach that can run on most consumer-level \glspl{gpu}. As a result, our method is particularly suitable for commercial detection devices by limiting their production costs. The implementations to reproduce our results and testing are available at \url{https://github.com/ChristophReich1996/3D_Baggage_Segmentation}.

\section{Related Work}
\begin{figure}[t]
    \centering
    \begin{subfigure}[b]{0.2\columnwidth}
        \centering
        \includegraphics[]{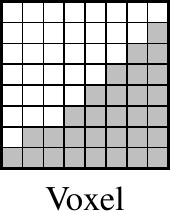}
        \label{fig:voxel}
    \end{subfigure}\hspace{0.3cm}
    \begin{subfigure}[b]{0.2\columnwidth}
        \centering
        \includegraphics[]{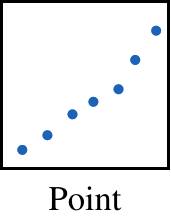}
        \label{fig:point}
    \end{subfigure}\hspace{0.3cm}
    \begin{subfigure}[b]{0.2\columnwidth}
        \centering
        \includegraphics[]{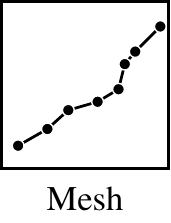}
        \label{fig:mesh}
    \end{subfigure}\hspace{0.3cm}
    \begin{subfigure}[b]{0.2\columnwidth}
        \centering
        \includegraphics[]{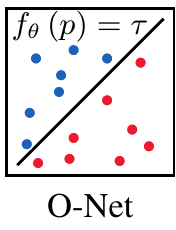}
        \label{fig:classifierspace}
    \end{subfigure}
    \caption{Various 3D representations (simplified 2D variants)~\cite{Mescheder2019}.}
    \label{fig:3drep}
\end{figure}

The methods and architectures used to perform 3D semantic segmentation depend on the representation of 3D data. In the following, we briefly introduce data types and corresponding architectures.\\
\indent A common representation of 3D data is point clouds (\textit{cf}.~\cref{fig:point}), which can be recorded using \gls{lidar} scanners. Points are recorded sparsely and distributed in space leading. This results in only the points of shapes stored with the background left out. Effective architectures used to segment point clouds are PointNet~\cite{Qi2017} and its improved successor PointNet++~\cite{Qi2017a}. We note that due to the lack of neighborhood information, convolutional methods are unfeasible. Due to this, point clouds are often converted to voxelized volumes~\cite{Tchapmi2018,Qi2016}.\\
\indent Meshes use multiple polygons to represent shapes (\textit{cf}.~\cref{fig:mesh}). An often used mesh type is the triangular mesh named after the triangles used to represent shapes. More precisely, meshes are similar to a graph spanned by vertices and edges. Vertices directly correspond to points in point clouds. Due to this connection to point clouds, meshes suffer similar disadvantages. An advantage, however, is the additional neighborhood information between vertices. Meshes have been used in 3D semantic segmentation tasks in combination with \glspl{cnn} or graph convolutions~\cite{WANG2018128,Kan2015,Masci2015,Taime2018}.
Voxel representation can be described as a cubical volume divided into a regular grid (\textit{cf}.~\cref{fig:voxel}). Each sub-cube in this regular grid is called a voxel. In contrast to point clouds, voxels are dense, \textit{i.e.}, the background is part of the grid. Each voxel representing a shape takes a value that describes material information. Background voxels are all associated with a single value, too. An advantage of this type of representation is that \glspl{cnn} with 3D convolutions can be applied, replacing a manual feature extraction with an automated one. One disadvantage over point clouds is the loss of detail that occurs depending on the resolution. With the background also being part of the representation, memory consumption also increases. To counter the described issues, attempts like reducing the input resolution or extracting features from different resolutions are made~\cite{Chen2016,Wang2019}.\\
\indent Liu et al.~\cite{Liu2019} present another approach combining the advantages of both voxels and point clouds. The key idea is to limit the operations on a voxel representation when necessary and otherwise use a point cloud representation instead. This way, a smaller memory footprint is maintained while also exploiting data locality and regularity.\\
\indent Given that at this point, the research community did not find a universally applicable \gls{cnn}-approach for large 3D volumes, they step back to 2D. This way, advantages of \glspl{cnn} can be used to an extent without running into memory problems like 3D \glspl{cnn} would do.\\
\indent The multi-view representation splits up the 3D volume into 2D image slices~\cite{Qi2016,Feng2018,SnapNet}. As a result, \glspl{cnn} can be applied but are restricted to 2D convolution. However, some global context is lost due to the split among the chosen axis.\\
\indent Because all mentioned representations suffer from a tradeoff between high memory consumption, detail, and applicable architectures, attempts are made to optimize such. One approach is to use the octree~\cite{Wang2017} data structure to represent the shapes in a tree-based format. Another approach introduces the \gls{onet}~\cite{Mescheder2019} that utilizes the function space of a neural network to model shapes (\textit{cf}.~\cref{fig:classifierspace}). The basic idea of \glspl{onet} is to represent a shape as a neural network classifier. For any point of interest, the classifier can predict whether or not the point is part of the shape. The \gls{onet} is versatile when it comes to input data. For example, one can transfer points clouds, single images from a multi-view representation, and voxels into a classifier setting.\\
\indent A neural network architecture commonly used for semantic segmentation is the U-Net~\cite{Ronneberger2015} and its adaption to the 3D U-Net~\cite{Cicek2016}. U-Net networks consist of an equal amount of down- and upsampling steps giving it the iconic U shape. The upsampling steps also take the output of the downsampling step at the same level as input. This residual connection enables the upsampling steps to take into account features determined by the downsampling steps. With the U-Net only being able to process 2D images, the 3D U-Net extends its functionality to a 3D input by replacing all 2D components with its 3D counterparts.

\section{Method}\label{sec:method}
\subsection{Dataset 3D-CT}
Our available dataset includes 2925 labeled CT scans of baggage containing firearms. For a better generalization, prohibited objects reach from ammo to pistols and even submachine guns. The CT scans are represented as voxelized volumes where each voxel contains material information.\\
\indent The further analysis involves semantic segmentation of the objects instead of a simple detection. For this purpose, the labels are given as a voxel representation, too. Each voxel indicates whether or not it is part of a firearm shape. Hence, the semantic segmentation task is binary. There is only one shape per instance.\\
\indent The CT scans have a resolution of $n \times 416 \times 616$ where $n$ varies among scans. To produce unified dimensions, $n$ is capped or padded to a value of $640$. The final dimensions of $640 \times 416 \times 616$ translate to a memory consumption of approx. $650$ Megabytes (MB) per scan.\\
\indent Each material value is further normalized to fit a range of 0 to 1.
After the preprocessing, the dataset is split into a training set of 2600 samples, a validation set of 128 samples, and a testing set of 196 samples. We further refer to this transformed dataset as \textit{3D-CT}.

\subsection{Super Resolution Occupancy Network for Semantic Segmentation} \label{subsec:onetss}

\begin{figure}[t]
    \centering
    \includegraphics[]{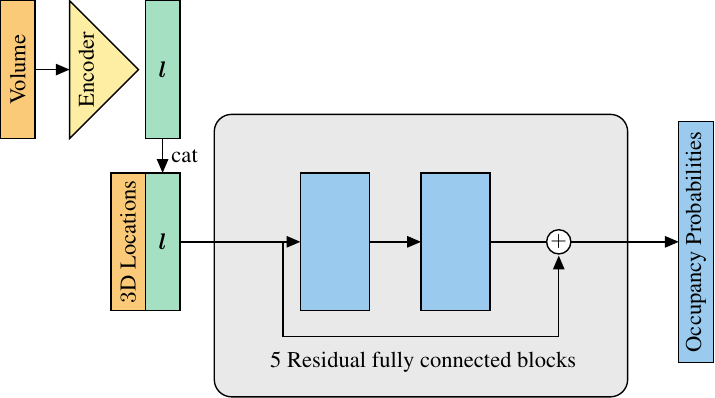}
    \caption{Alternative super-resolution \gls{onet} architecture.}
    \label{fig:O_net_cat}
\end{figure}

The general idea of \gls{onet} is to define a 3D shape as a classifier for $\mathbb{R}^3$. More precisely, each 3D coordinate gets mapped onto a pseudo probability which is called occupancy value. Occupancy values give an estimation of whether or not a coordinate is part of the shape. Using a threshold value this decision can be influenced manually. Formally, the classification task can be described as
\begin{equation}
    f_{\theta}:\mathbb{R}^3\times\Tensor{L}\to\left[0, 1\right],
\end{equation}
where $l \in \Tensor{L}$ is a finite observation of $\mathbb{R}^3$ and $\theta$ are the parameters of $f$. \glspl{onet} generate $l$ by using an arbitrary neural network architecture to encode a shape. The mapping $f$ itself is a fully connected decoder neural network. The input to the encoder depends on the chosen architecture. For voxelized volumes, a \gls{cnn} encoder is used. The decoder parameters $\theta$ are trained by randomly sampling 3D coordinates from the \glspl{bb} of targeted shapes. Methods exist to transform the classifier representation of a shape back into a mesh representation~\cite{Mescheder2019}. Nonetheless, a significant speedup is possible if test and validation metrics are also calculated on sampled coordinates.\\
\indent We use the \gls{onet} to perform 3D semantic segmentation on voxelized volumes. In particular, we focus on solving the semantic segmentation of the 3D-CT dataset. Since the dataset results in a binary semantic segmentation task, this task is equal to defining 3D shapes, making the \gls{onet} a suitable architecture.\\
\indent For the 3D-CT dataset, a single volume consumes approx. $650\text{MB}$. Fortunately, \glspl{onet} reduce the memory consumption for the decoder in comparison to U-Nets, as complete volumes do not have to be processed simultaneously. Nevertheless, super-resolution is needed to manage the memory footprint of the inputted volumes that are processed by the \gls{cnn} encoder. For the super-resolution \gls{onet}, volumes are downsampled using average pooling before they are encoded even though the network classifies the high-resolution coordinates. In other words, training is performed by sampling high-resolution coordinates with their corresponding high-resolution labels while encoding low-resolution volumes. For the inference of a complete volume, the \gls{mise} algorithm~\cite{Mescheder2019} is used but stopped before conducting the Marching Cubes algorithm. Originally, Mescheder et al.~\cite{Mescheder2019} use a CPU version of \gls{mise}. To bypass this restriction, we provide an implementation of a \gls{gpu} variant applicable to voxelized volumes.\\
\begin{figure}[t]
    \centering
    \includegraphics[]{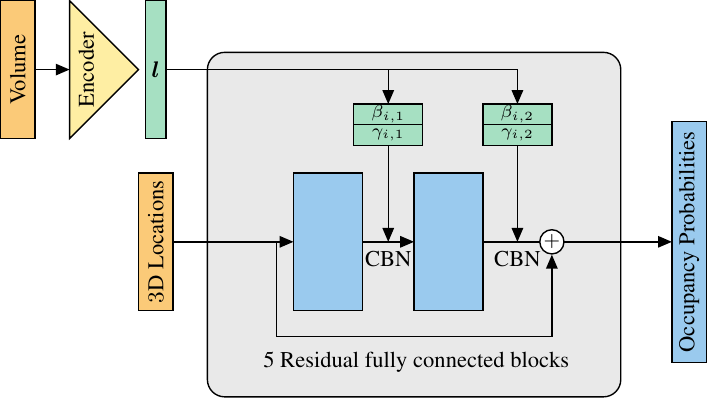}
    \caption{Super-resolution \gls{onet} architecture.}
    \label{fig:O_net}
\end{figure}
The originally proposed \gls{onet} uses \gls{cbn}~\cite{cbatchnorm} to pass the output of the encoder to the decoder. In traditional batch normalization~\cite{batchnorm}, a layer with $d$-dimensional input $\Tensor{X}=(X^{(1)}...X^{(d)})$ is normalized by
\begin{equation}
    \gamma^{(c)}\frac{\Tensor{X}^{(c)}-\Mean{\Tensor{X}^{(c)}}}{\sqrt{\Var{\Tensor{X}^{(c)}}}}+\beta^{(c)}
\end{equation}
where $\gamma_c$ and $\beta_c$ are learnable parameters per channel $c$. In \gls{cbn}, however, $\gamma(x)^{(c)}$ and $\beta(x)^{(c)}$ are learnable mappings of some input $x$. In \glspl{onet}, $x=l$ holds for any \gls{cbn} layer. The previously mentioned mappings are typically implemented as fully connected layers. As a consequence, encodings are only inserted into an \gls{onet} as normalization parameters (\textit{cf}.~\cref{fig:O_net}). For this reason, we propose an alternative \gls{onet} architecture (\textit{cf}.~\cref{fig:O_net_cat}). This architecture uses a concatenation of the coordinates with their corresponding (repeated) latent vectors as the decoder input.

\subsection{Occupancy Network for Bounding Boxes}

The output of an \gls{onet} can also be used to build an axis-aligned \gls{bb} around a gun. If a \gls{bb} can be reliably constructed, it is sufficient to perform semantic segmentation only within the \gls{bb}. For this adaptation, the basic \gls{onet} architecture described in \cref{subsec:onetss} stays untouched. Nonetheless, in addition to using low-resolution input volumes, only low-resolution coordinates and labels are sampled. Low-resolution labels are defined by max-pooling associated high-resolution labels. A \gls{bb} is simply constructed by using the lowest and highest predicted coordinates.
For this use case, an \gls{onet} is not required to perform super-resolution or to extract fine details which represents a much easier task than performing super-resolution 3D semantic segmentation.

\subsection{HiLo-Network}

The challenge for a super-resolution deep neural network is retrieving high-resolution information from low-resolution representations. In our approach, the HiLo-Network, we circumvent this complex generative task by inputting dense high-resolution information into a neural network. However, the main bottleneck, \textit{i.e.}, \gls{gpu}-acceleration of gradient-based optimization, is improved by applying a divide and conquer procedure to semantic segmentation. During gradient descent, only small chunks of each instance within a batch need to be loaded into \gls{vram}.\\
\indent Independent of a particular architecture, the main concept of inference for HiLo-Networks is as follows. Instead of passing a complete volume into a network, only a small 3D chunk (window) of fixed-size $w^3$ is used. One can feed multiple windows into the network to conduct semantic segmentation on the full volume. Nonetheless, global relations between different windows are not taken into consideration yet. To overcome this problem, a common computer vision trick is applied. Centered around the first window, a seconder bigger window of size $(w * d)^3$ is constructed but downsampled by a factor $d$ using average pooling. This small pyramid contains high-resolution local information as well as low-resolution global information. If a window crosses the borders of a volume, zero padding is applied. In our tests for small values of $w$, the pooling operation is conducted on the CPU without a significant increase in computation time. In case two windows are not sufficient, one can easily add more levels. The memory consumption of the fixed-size pyramid grows linear in terms of levels. In comparison, memory consumption grows nonlinear in the size of the volume dimensions. The downside of our approach is that multiple HiLo-Network passes are needed to segment a complete volume. However, in a likely real-time scenario in which only a few volumes per time step are processed, multiple pyramids per volume can be passed simultaneously.


\begin{figure}[t]
    \centering
    \includegraphics[]{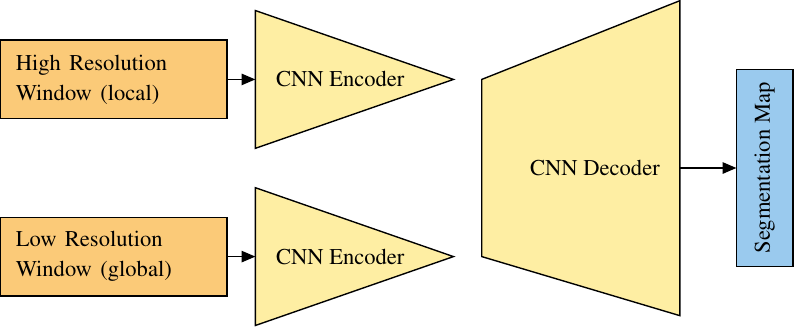}
    \caption{Architecture of the \gls{cnn}-approach.}
    \label{coarse_cnn_fig}
\end{figure}

\begin{figure}[t]
    \centering
    \includegraphics[]{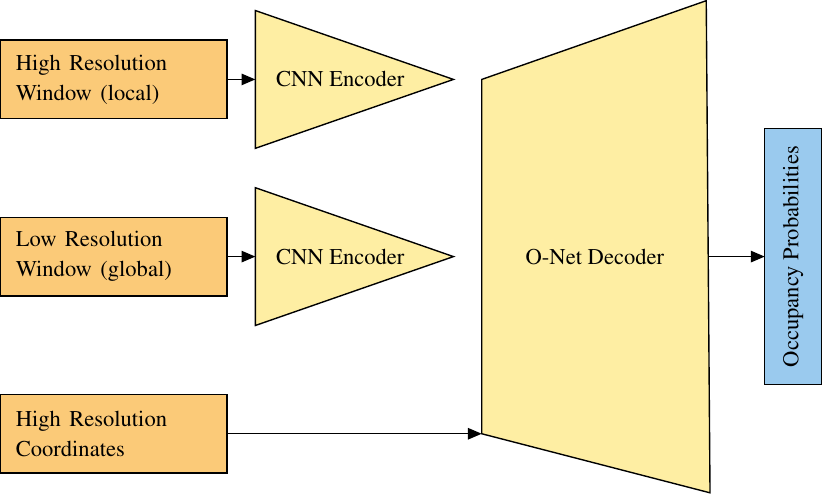}
    \caption{Architecture of the \gls{onet}-approach.}
    \label{coarse_onet_fig}
\end{figure}

Coarse sketches of the utilized network architectures are displayed in \cref{coarse_cnn_fig} and \cref{coarse_onet_fig}. The first one consists of a \gls{cnn} encoder and an \gls{onet} decoder~\cite{Mescheder2019} (\gls{onet}-approach). Because small 3D windows can be handled by consumer \glspl{gpu}, the second architecture includes a \gls{cnn} encoder as well as a \gls{cnn} decoder (\gls{cnn}-approach). In both architectures, we train separate encoders per pyramid level. The resulting encodings are concatenated before they are processed by the decoder. For the \gls{onet}-approach, a query of 3D coordinates has to be fed into the decoder, too.\\
\indent Both architectures no longer need to be trained on complete instances but rather on pyramids constructed at randomly drawn locations within each instance. Any two pyramids with different locations but of the same instance can be independently processed. As a result, the sampling scheme of the pyramid locations can be biased. This comes in handy, as the considered CT scans mostly contain air. Detecting air is easy since it is related to only a single material value. Hence, regularly redrawing locations at which the surrounding pyramid does not contain a gun can reduce training time. More precisely, at each training step, a random location from each instance inb] a batch is drawn and, if needed, redrawn with a likelihood of 90\%. Afterward, pyramids are built centered at those locations and processed by the HiLo-Network. For parameter validation a similar logic holds. Each time the validation set is evaluated, only one pyramid per instance is processed, too. Estimating validation losses leads to a significant decrease in training time. The estimation can be improved by considering a moving average over validation losses. Nevertheless, testing results can only be determined by evaluating non-overlapping\footnote{Non-overlapping on the first pyramid level.} pyramids that fully cover a test instance.

\subsection{Training Queues}
Besides \gls{vram} management during gradient descent optimization, data loading is a major concern. Although only pyramids with a comparably small memory footprint have to be loaded on the \gls{gpu}, full instances have to be processed on the host. For quick access, files have to be stored in the host RAM. However, loading a dataset build of a few thousand large files into host RAM is not feasible. Common deep learning libraries provide multi-threading mechanisms to load files into host RAM parallel to performing gradient descent on the \gls{gpu}. This allows to partially circumvent slow disk access by computations. Nonetheless, multiple files have to be loaded to form a batch and, by construction, one thread/core can only load a single file into host RAM. Therefore, these mechanisms are only useful on expensive workstations that are not jointly used by other tasks.\\
\indent We exploit the fact that host RAM is usually bigger and cheaper than \gls{vram}. For this, a training queue of a user-defined size is constructed in host RAM. At each training step, a batch is randomly drawn from the training queue rather than loaded from the hard drive. Before each batch formation, one new file is loaded from the disk and added to the training queue. Likewise, one instance is deleted from the training queue if the user-defined queue size is reached. The training queue procedure reduces the number of files loaded from the hard drive to one per processed batch. The loading happens parallel to gradient descent. If the queue size is sufficient, the data distribution can still be represented well enough.\\
\indent We inspect three different training queues that vary in the implementation of sampling, adding, and deleting. First, a queue that uniformly draws instances and replaces the instance that has been inserted first, \textit{i.e.} \textit{\gls{fifo}}. Second, a queue weighing an instance $i$ according to its number of occurrences $o_i$ (occurrence based). The more often an instance has already been drawn, the less the probability to draw it again. Formally, the probability that $i$ is drawn equals $e^{-o_i} / \sum_{j} e^{-o_j}$ and the instance with the most occurrences is replaced. Third, a queue that weighs instances according to their hardness (hardness based). The worse the training loss $h_i$ of an instance $i$, the bigger its probability to get drawn again. The instance with the best training loss is replaced. Formally, the probability that $i$ is drawn equals $e^{h_i} / \sum_{j} e^{h_j}$.

\section{Experiments}

\subsection{Super-resolution Occupancy Network}\label{superresresults}
Within this section, we evaluate and compare the performance of the proposed super-resolution \gls{onet} architectures for semantic segmentation on the 3D-CT dataset. At this, we describe in detail model architectures, training processes, and implementations.\\
\indent The \gls{cnn} encoders of the super-resolution \glspl{onet} are built from five standard convolutional residual blocks~\cite{resnet}. After the five residual blocks, a final fully connected layer is used to produce a scalar occupancy value per inputted coordinate. Volumes inputted to the encoder are downsampled by a factor of 8 using average pooling. We test a wide and a shallow \gls{cnn} encoder. The wide one utilizes twice as many filters as the shallow one per layer. Super-resolution \gls{onet} decoders are implemented with five fully connected residual blocks.\\
\indent For any but the last layer of a network, a leaky ReLU~\cite{leakyrelu} is utilized as the activation function in combination with a normalization operation. Output layers apply a sigmoid activation function. Depending on the tested architecture, batch normalization or \gls{cbn} is deployed.  If \gls{cbn} is used, the latent vector of the encoder is processed by two leaky ReLU activated fully connected layers to predict the normalization parameters. To optimize the \gls{onet} parameters the Adam optimizer~\cite{adam} is used. Adam is initialized with a learning rate of $0.001$ as well as PyTorch default parameters and optimizes a binary cross-entropy loss. The dataset is processed in batches of size 8. Each network is trained for 200 epochs, which takes approximately 12h on a single \gls{gpu}. Validation is conducted after every epoch.\\
\indent At training time, $2^{14}$ coordinates are sampled per instance and training step. For the 3D-CT dataset, the class labels are highly unbalanced. We cope with this problem by using a biased sampling scheme. The applied scheme draws 60\% of the coordinates uniformly from the targeted shapes. The remaining coordinates are sampled uniformly over the whole volume. A gun classification is assumed in case the output of a network is $>0.5$.\\
\indent Our tests differ in concatenation (Cat) and no concatenation, \gls{cbn}, and batch normalization as well as the wide encoder (Wide) and shallow encoder. To compare the results of the different architectures, the \gls{iou} is computed on a 3D-CT test set. During testing, we sample $2^{18}$ coordinates, uniformly drawn from each full volume, for a sophisticated approximation of the \gls{iou}.

\begin{table}[t]
    \centering
    \begin{center}
        \begin{tabular}{ccccc}
            \toprule
            Cat & \gls{cbn} & Wide & Parameters & \gls{iou} \\ 
            \midrule 
            \cmark & \cmark & \cmark & 3.4$\si{\mega\nothing\nothing}$ & 0.1744 \\ 
            \cmark & \xmark & \cmark & 2.2$\si{\mega\nothing}$ & 0.1591 \\ 
            \xmark & \cmark & \cmark & 3.3$\si{\mega\nothing}$ & 0.1689 \\ 
            \cmark & \cmark & \xmark & 2.0$\si{\mega\nothing}$ & {\bf 0.1943} \\ 
            \cmark & \xmark & \xmark & 0.7$\si{\mega\nothing}$ & 0.1612 \\ 
            \xmark & \cmark & \xmark & 1.9$\si{\mega\nothing}$ & 0.1128 \\ 
            \bottomrule
        \end{tabular} 
    \end{center}
    \caption{Test results of different architecture choices for the super-resolution \gls{onet}.}
    \label{tab:onet_results_low_high}
\end{table}

From the low \gls{iou} scores, shown in \cref{tab:onet_results_low_high}, we conclude that no \gls{onet} architecture is able to perform super-resolution and 3D semantic segmentation at once. Due to this observation, we decide to evaluate the same models directly on low-resolution labels to test the general 3D semantic segmentation capabilities of \glspl{onet}. The low-resolution labels are downsampled by a factor of 8 using max pooling.

\begin{table}[t]
    \centering
    \begin{center}
        \begin{tabular}{ccccc}
            \toprule
            Cat & \gls{cbn} & Wide & Parameters & \gls{iou} \\ 
            \midrule
            \cmark & \cmark & \cmark & 3.4$\si{\mega\nothing}$ & 0.1843 \\ 
            \cmark & \xmark & \cmark & 2.2$\si{\mega\nothing}$ & {\bf 0.1968} \\ 
            \xmark & \cmark & \cmark & 3.3$\si{\mega\nothing}$ & 0.1554 \\ 
            \cmark & \cmark & \xmark & 2.0$\si{\mega\nothing}$ & 0.1836 \\ 
            \cmark & \xmark & \xmark & 0.7$\si{\mega\nothing}$ & 0.1805 \\ 
            \xmark & \cmark & \xmark & 1.9$\si{\mega\nothing}$ & 0.1643 \\ 
            \bottomrule 
        \end{tabular} 
    \end{center}
    \caption{Test results of different \gls{onet} architectures for a \textit{low-resolution input volume and low-resolution coordinates}.}
    \label{tab:onet_results_low_low}
\end{table}

\begin{figure}[t]
    \centering
    \includegraphics{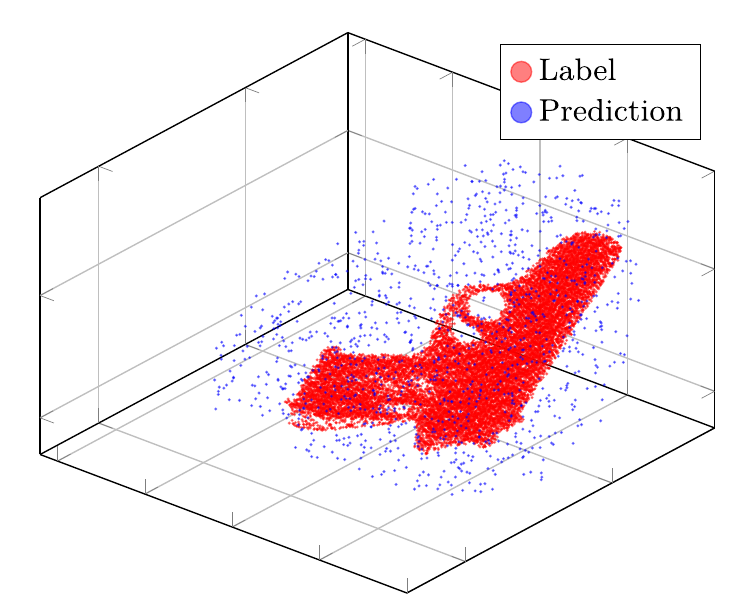}
    \caption{Exemplary super resolution \gls{onet} semantic segmentation result.}
    \label{fig:onet_result_1}
\end{figure}

\begin{figure}[!ht]
    \centering
    \includegraphics{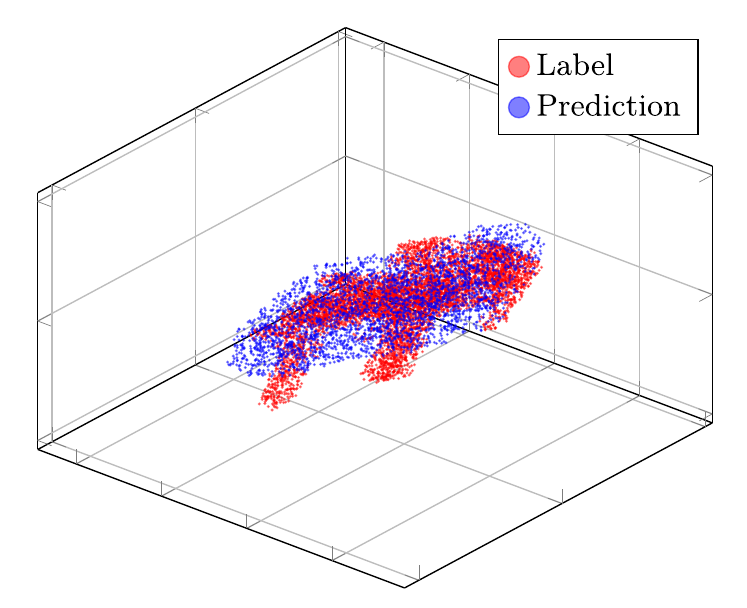}
    \caption{Exemplary super resolution \gls{onet} semantic segmentation result.}
    \label{fig:onet_result_2}
\end{figure}

\begin{figure}[!ht]
    \centering
    \setlength{\figW}{\columnwidth}
    \setlength{\figH}{0.6\columnwidth}
    \includegraphics[]{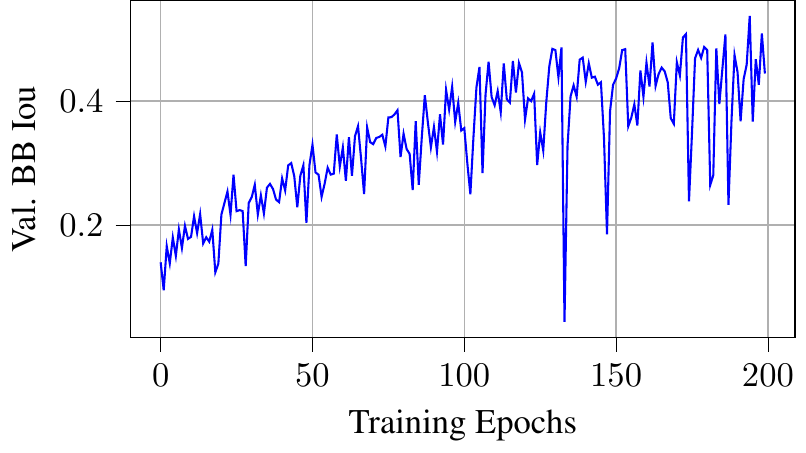}
    \caption{Curve of validation \gls{iou}s during training of our alternative \gls{onet} architecture.}
    \label{fig:val_bb_iou_no_cbn}
\end{figure}

\begin{figure}[!ht]
    \centering
    \setlength{\figW}{\columnwidth}
    \setlength{\figH}{0.6\columnwidth}
    \includegraphics[]{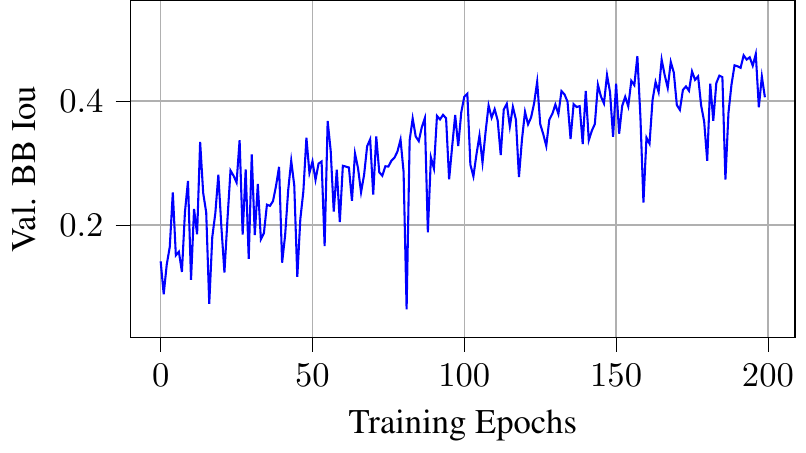}
    \caption{Curve of validation \gls{iou}s during training of the standard \gls{onet} architecture.}
    \label{fig:val_bb_iou_cbn}
\end{figure}

We conclude from the results shown in \cref{tab:onet_results_low_low}, that \glspl{onet} in general is not able to segment firearms properly. We analyze the segmentation results visually to get an understanding of this failure.
From these visualizations (\textit{cf}.~\cref{fig:onet_result_1} and \cref{fig:onet_result_2}), we can observe that \glspl{onet} can predict the rough position of an object accurately. However, the \gls{onet} is not capable of predicting detailed shapes. The described behavior leads us to the idea of using \glspl{onet} to predict a \gls{bb} rather than a fine segmentation.\\
\indent We can also observe that the whole training process tends to be noisy since the validation metrics have a high fluctuation (\textit{cf}.~\cref{fig:val_bb_iou_no_cbn} and \cref{fig:val_bb_iou_cbn}). This is most likely due to the random nature of \glspl{onet}.

\subsection{Occupancy Network for Bounding Boxes}

We test the same networks as in \cref{superresresults} for \gls{bb} construction. Before the construction of \glspl{bb}, a margin of 10 is subtracted from and added to the minimum and maximum predicted coordinates respectively. This procedure improves \gls{bb} \gls{iou} results. 

\begin{table}[t]
    \centering
    \begin{center}
        \begin{tabular}{ccccc}
            \toprule
            Cat & \gls{cbn} & Wide & Parameters & \gls{bb} \gls{iou} \\ 
            \midrule 
            \cmark & \cmark & \cmark & 3.4$\si{\mega\nothing}$ & 0.5093 \\ 
            \cmark & \xmark & \cmark & 2.2$\si{\mega\nothing}$ & 0.5056 \\ 
            \xmark & \cmark & \cmark & 3.3$\si{\mega\nothing}$ & 0.4162 \\ 
            \cmark & \cmark & \xmark & 2.0$\si{\mega\nothing}$ & {\bf 0.5314} \\ 
            \cmark & \xmark & \xmark & 0.7$\si{\mega\nothing}$ & 0.5039 \\ 
            \xmark & \cmark & \xmark & 1.9$\si{\mega\nothing}$ & 0.3833  \\ 
            \bottomrule 
        \end{tabular} 
    \end{center}
    \caption{Test results of different \gls{onet} architectures (low-resolution input volumes, high-resolution coordinates) for predicting an axis-aligned \gls{bb}.}
    \label{tab:onet_results_low_high_bb}
\end{table}

\begin{table}[t]
    \centering
    \begin{center}
        \begin{tabular}{ccccc}
            \toprule
            Cat & \gls{cbn} & Wide & Parameters & \gls{bb} \gls{iou} \\ 
            \midrule 
            \cmark & \cmark & \cmark & 3.4$\si{\mega\nothing}$ & 0.5191 \\ 
            \cmark & \xmark & \cmark & 2.2$\si{\mega\nothing}$ & {\bf 0.5640} \\ 
            \xmark & \cmark & \cmark & 3.3$\si{\mega\nothing}$ & 0.4522 \\ 
            \cmark & \cmark & \xmark & 2.0$\si{\mega\nothing}$ & 0.5118 \\ 
            \cmark & \xmark & \xmark & 0.7$\si{\mega\nothing}$ & 0.5269 \\ 
            \xmark & \cmark & \xmark & 1.9$\si{\mega\nothing}$ & 0.4710 \\ 
            \bottomrule 
        \end{tabular} 
    \end{center}
    \caption{Test results of different \gls{onet} architectures (low-resolution input volumes, low-resolution coordinates) for predicting an axis-aligned \gls{bb}}
    \label{tab:onet_results_low_low_bb}
\end{table}

The test results shown in \cref{tab:onet_results_low_high_bb} and \cref{tab:onet_results_low_low_bb} confirm the observed behavior that \glspl{onet} are able to detect the coarse region around a shape. Additionally, a slight performance increase can be observed when training on a low-resolution label rather than on a high-resolution label.\\
\indent Furthermore, it can be derived from the results that \gls{cbn} performs worse than our alternative \gls{onet} architecture. Since \gls{cbn} uses more parameters, too, it can be recommended to refrain from using it for \gls{bb} construction.\\
\indent We also conclude from all \gls{onet} experiments that rather the \gls{onet} decoder is the main performance bottleneck. 
We justify this by the minor influence of the shallowness of the encoder on the overall performance.

\subsection{HiLo-Network}
\begin{table}[t]
    \begin{center}
        \begin{tabular}{c|ccc}
        \toprule
         & $w=16$ & $w=32$ & $w=48$  \\
         \midrule
         $d=2$   & 0.6216 & \textbf{0.6836} & \textbf{0.7059} \\
         $d=4$   & 0.6168 & 0.6700 & 0.6970 \\
         $d=8$   & \textbf{0.6488} & 0.6640 & 0.6233 \\
         \bottomrule
        \end{tabular}
    \end{center}
    \caption{Test \gls{iou}s for the \gls{cnn}-approach. The results are broken down into window size $w$ and downsampling factor $d$.}
    \label{unetresults}
\end{table}

\begin{table}[t]
    \begin{center}
        \begin{tabular}{cc}
        \toprule
        window size $w$ & \gls{vram} \\
        \midrule
        $16$ & $1.1$ GB \\
        $32$ & $4.5$ GB \\
        $48$ & $12.3$ GB \\
        \bottomrule
        \end{tabular}
    \end{center}
    \caption{\gls{vram} consumption of different window sizes $w$ denoted in Gigabytes (GB) during training with batch size 16 and our \gls{cnn}-approach.}
    \label{ramresults}
\end{table}

\begin{table}[t]
    \begin{center}
        \begin{tabular}{c|cccc}
        \toprule
         & $w=16$ & $w=32$ & $w=48$ &  \\
         \midrule
         $d=2$   & 0.4248 & 0.5395 &  \textbf{0.5360} &  \\
         $d=4$   & \textbf{0.4481} & \textbf{0.5962} &  0.4992 &  \\
         $d=8$   & 0.3963 & 0.4953 &  0.4400 &  \\
        \bottomrule
        \end{tabular}
    \end{center}
    \caption{Test \gls{iou}s for the \gls{onet}-approach with window size $w$ and downsampling factor $d$.}
    \label{onetresults}
\end{table}

\begin{figure}[!ht]
    \centering
    \includegraphics{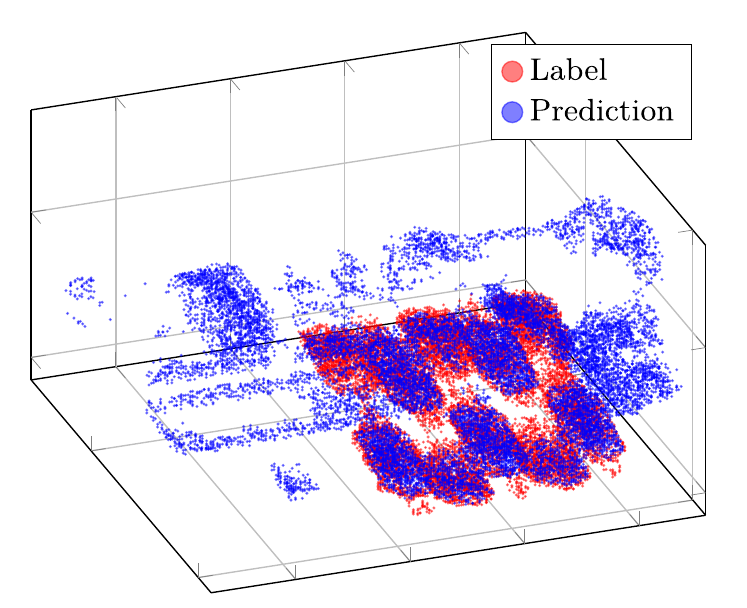}
    \caption{Exemplary HiLo-Network \gls{cnn}-approach semantic segmentation result.}
    \label{fig:hilo_result_2}
\end{figure}

In this section, we evaluate HiLo-Networks on the 3D-CT dataset and give a detailed description of training procedures as well as architectures. In particular, we evaluate the two main hyperparameters of HiLo-Networks, the window size $w$, and the downsampling factor $d$, in terms of test \gls{iou} and \gls{vram} consumption. Furthermore, the \gls{cnn}-approach is compared to the \gls{onet}-approach.\\
\indent We implement all HiLo-Network encoders with 6 residual blocks~\cite{resnet}. Due to the small window sizes, only one average pooling operation is conducted after the first block in the \gls{cnn}-approach. The decoder of the \gls{cnn}-approach consists of 7 residual blocks with an upsampling operation after the first block. The decoder of the \gls{onet}-approach uses 3 fully connected residual blocks. We use our alternative \gls{onet} decoder architecture. In contrast to the \gls{cnn}-approach, an average pooling operation is conducted after the first, second (for $w=16, w= 32, w=48$), third (for $w= 32, w=48$), and fourth (for $w=48$) encoding block. This prevents the first fully connected layer of the \gls{onet}-approach decoder to grow untrainable large.\\
\indent For any but the last layer of a network, SELU~\cite{selu} is used as the non-linearity in combination with batch normalization~\cite{batchnorm}. Output layers deploy a sigmoid activation function. All networks are trained using the Adam optimizer~\cite{adam}. The optimizer is initialized with a learning rate of $0.001$ and PyTorch default parameters. The dataset is processed in batches of size 16. For the \gls{cnn}-approach, the focal loss~\cite{focal} whereas for the \gls{onet}-approach the binary cross-entropy is minimized. The focal loss manages unbalanced class occurrences automatically. In early testings, the focal loss appeared to be not useful for the \gls{onet}-approach. Likely due to the random nature of \glspl{onet}. The network parameters are validated every $128$ training steps. The parameter set with the best validation \gls{iou} is used for testing. Each network is trained for $20$ epochs. Training queues of the first type (\gls{fifo}) and of size 512 are used. The \gls{onet}-approach is trained and tested by sampling $2^{11}$ 3D coordinates per query. Since the considered windows are significantly smaller than the volumes from previous approaches, fewer sampled points are used. The threshold for HiLo-Network classifying a gun is set to the output $>0.5$.\\
\indent In sum, 18 networks are trained to evaluate the HiLo-Network. On average, training of one HiLo-Network takes 3 days on a single \gls{gpu}. In addition, many more networks were trained during the development of HiLo-Networks. Unfortunately, not only memory consumption of volumes grows non-linear in terms of dimension sizes but also the inference runtime of HiLo-Networks. As we need to be responsibly wth computational resources, we make simplifying assumptions for training, validating, and testing. For training and validating, pyramid locations are only sampled from the axis-aligned \glspl{bb} of firearms. Additionally, we only load \glspl{bb} instead of complete files into host RAM. For testing, we follow a similar approach. However, to strengthen our results we extend the \glspl{bb} by 50 voxels in each direction. This procedure is legitimized by the construction of the 3D-CT dataset. The dataset is created artificially with a limited number of objects. For each gun, there are multiple instances for which only the spatial arrangement of the objects varies. Hence, most object-gun combinations are covered by extended \glspl{bb}. Furthermore, huge parts of 3D-CT volumes represent air. \\
\begin{figure}[t]
    \centering
    \includegraphics{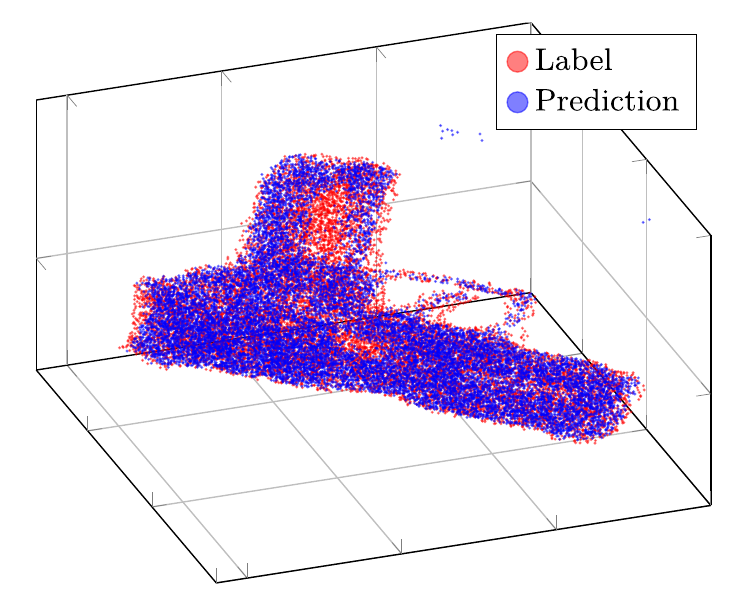}
    \caption{Exemplary HiLo-Network \gls{cnn}-approach semantic segmentation result.}
    \label{fig:hilo_result_1}
\end{figure}
The test \gls{iou}s for the \gls{cnn}-approach can be seen in \cref{unetresults}. For the small window size $w=16$, the biggest tested downsampling factor of $d=8$ results in a strong \gls{iou} improvement in comparison to $d=2$ or $d=4$. For $w=32$ and $w=48$ we observe a contrasting behavior. We account this to the specifications of the 3D-CT dataset. In particular, we assume that material properties are more important than shapes. As a consequence, a \textit{huge} moving window instead of a moving pyramid seems to be already sufficient for 3D semantic segmentation. Surprisingly, the \gls{iou} seems to decrease if the downsampling factor is increased. We apply zero-padding in case a pyramid level is partially out of range for a volume. We presume that this leads to worse results. As the downsampling factor is increased, this effect is amplified in theory and describes the observed behavior. In \cref{ramresults} the \gls{vram} consumption during training time is displayed. It can be seen that using window sizes of $w=16$ or $w=32$ allows to train HiLo-Networks even on consumer \glspl{gpu} while mostly preserving test \gls{iou} results. For visualizations of \gls{cnn}-approach results see \cref{fig:hilo_result_1} and \cref{fig:hilo_result_2}.\\
\indent The results for the \gls{onet}-approach, shown in \cref{onetresults}, are not discussed in detail. Overall, test \gls{iou}s are significantly worse than those of the \gls{cnn}-approach and highly unstable. This indicates a general problem of \glspl{onet} for 3D semantic segmentation, at least in combination with a sliding pyramid/window. Hence, variations due to varying values of $w$ and $d$ cannot be reasonably explained.

\subsection{Training Queues}

\begin{table}[t]
    \begin{center}
        \begin{tabular}{cc}
        \toprule
            & \gls{iou} \\
            \midrule
          \gls{fifo} & 0.6836 \\
          Hardness & 0.6615 \\
          Occurrence & \textbf{0.6838} \\
        \bottomrule
        \end{tabular}
    \end{center}
    \caption{Test results for the \gls{cnn}-approach with different training queues.}
    \label{queueresults}
\end{table}

In \cref{queueresults}, test \gls{iou}s after 20 training epochs are shown separately for the three introduced training queues. For this, a \gls{cnn}-approach HiLo-Network with $w=32$ and $d=2$ is trained. The size of each training queue is 512.\\
\indent Besides the general speed advantages of training queues, we cannot add additional benefits through more complex batch sampling schemes. We identify two reasons for this. Generally, independent of a particular gun, the surrounding objects are the same for any instance of 3D-CT. Hence, favoring instances over others barely improves training. The second reason only holds for the hardness-based queue. Pyramids/windows only represent small fractions of an instance and probably do not give a good estimate of the actual hardness of a complete instance.

\section{Conclusion}
In this work, we propose two novel architectures for 3D semantic segmentation of voxelized volumes. We test our methods on a 3D CT scan dataset for firearm detection in luggage. First, we introduce a super-resolution Occupancy Network.  Unfortunately, the tested super-resolution \gls{onet} architectures are not able to achieve the desired results. Nonetheless, we demonstrate that \glspl{onet} can be successfully deployed to extract \glspl{bb} of shapes. Second, we propose HiLo-Networks, a new scaleable neural network architecture to perform 3D semantic segmentation. Our experiments show that HiLo-Networks can reliably segment guns within the luggage. At this, they are memory efficient as well as scalable in terms of input resolution. In particular, HiLo-Networks can be trained on consumer-level \glspl{gpu}.\\
\indent Future research should evaluate HiLo-Networks on other datasets as we expect them to work on a broader range of 3D datasets. Another aspect to be investigated is the inference time. In our implementation full volumes are segmented by using multiple forward passes. Efficiently extracting a \gls{bb} before performing segmentation could be the key to reducing the inference time. For instance, our adaption of the \gls{onet} for \gls{bb} construction could be used. \\
\indent More recent work based on implicit neural representations include \cite{Peng2020, Sitzmann2020, Zhong2021, Reich2021, Mildenhall2020, Tewari2021} (as of November 2021).

\section*{Acknowledgements}
\noindent
We thank Professor Stefan Roth from the Visual Inference Lab at TU Darmstadt for the supervision and fruitful discussions. This work was part of the Project Lab Deep Learning in Computer Vision Winter Semester 2019/2020 at TU Darmstadt.

{\small
    \bibliographystyle{ieee_fullname}
    \bibliography{bib}
}

\section{Appendix}

\end{document}